\documentclass[11pt,a4paper]{article}
\usepackage[hyperref]{acl2017}
\usepackage{graphicx}
\usepackage{times}
\usepackage{latexsym}
\usepackage{tikz}
\usepackage{multirow}
\usepackage{pgf}
\usepackage{tikz-qtree}
\usepackage{CJK}
\usepackage{url}
\usepackage{amsmath}
\usepackage[version=3,arrows=pgf]{mhchem}
\usetikzlibrary{arrows,decorations,backgrounds,positioning,fit,shapes,calc}

\usepackage{pst-node}
\def\W#1#2{\rnode{#1}{#2}\hfill}

\DeclareMathOperator*{\argmax}{arg\,max}

\aclfinalcopy 

\title{Learning When to Attend for Neural Machine Translation}

\author{Junhui Li \\
  School of Computer Science and Technology \\
  Soochow University, Suzhou, China \\
  {\tt lijunhui@suda.edu.cn} \\\And
  Muhua Zhu \\
  Tencent AI Lab, Shenzhen, China \\
  {\tt muhuazhu@tencent.com} \\}

\date{}

\begin{document}

\begin{CJK}{UTF8}{gkai}

\maketitle
\begin{abstract}
  In the past few years, attention mechanisms have become an indispensable component of end-to-end neural machine translation models. However, previous attention models always refer to some source words when predicting a target word, which contradicts with the fact that some target words have no corresponding source words. Motivated by this observation, we propose a novel attention model that has the capability of determining when a decoder should attend to source words and when it should not. Experimental results on NIST Chinese-English translation tasks show that the new model achieves an improvement of 0.8 BLEU score over a state-of-the-art baseline.
\end{abstract}

\section{Introduction}
\begin{figure*}[t]
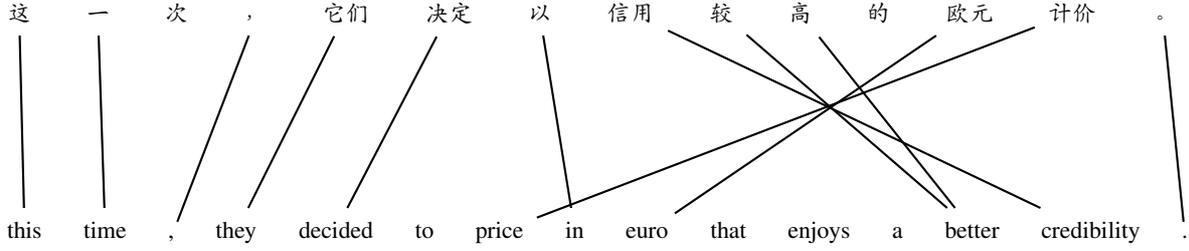

\small
\W{a}{这} \W{b}{一} \W{c}{次} \W{d}{，} \W{e}{它们} \W{f}{决定} \W{g}{以} \W{h}{信用} \W{i}{较} \W{j}{高} \W{k}{的} \W{l}{欧元} \W{m}{计价} \W{n}{。}

\vspace{2.5cm}
\W{A}{this} \W{B}{time} \W{C}{,} \W{D}{they} \W{E}{decided} \W{F}{to} \W{G}{price} \W{H}{in} \W{I}{euro} \W{J}{that} \W{K}{enjoys} \W{L}{a} \W{M}{better} \W{N}{credibility} \W{O}{.}

\psset{nodesep=5pt}
\ncline{a}{A} 
\ncline{b}{B}
\ncline{d}{C}
\ncline{e}{D}
\ncline{f}{E}
\ncline{g}{H}
\ncline{h}{N}
\ncline{i}{M}
\ncline{j}{M}
\ncline{l}{I}
\ncline{m}{G}
\ncline{n}{O}
\caption{An illustrating example of untranslated words, where the unaligned English words are not translated from any source (Chinese) words.} \label{unaligned_words}
\end{figure*}

The past several years have witnessed rapid progress of end-to-end neural machine translation (NMT) models, most of which are built on the base of encoder-decoder framework~\cite{Sutskeveretal:14,Bahdanauetal:15,Luongetal:15a}. In addition, attention mechanisms have become an indispensable component in state-of-the-art NMT systems~\cite{Luongetal:15b,Tuetal:16a}. The idea of attention mechanisms is to guide a translation decoder to selectively focus on a local window of source words that are used to generate the current target word. Previous studies have demonstrated necessity and effectiveness of such attention mechanisms.
\\
\indent However, previous attention models are all dedicated to solving the problem of where to attend. They take no account of when to attend. In fact, target words in a translation are not always generated according to the source sentence. Take the Chinese-English translation in Figure~\ref{unaligned_words} as an example, where words are manually aligned. The English words {\em to}, {\em enjoys}, {\em that}, {\em a} are not translated from any source words. Instead, it is appropriate to use a language model to predict the words by conditioning on their preceding words. To show how prevalent the phenomenon is, we analyze a set of 900 Chinese-English sentence pairs with manual word alignments~\cite{LiuSun:15}, and find that $25.3\%$  (21,244/28,433) English words are not translations from Chinese words. Thus, an attention mechanism should distinguish between target words that are generated referring to the source sentence and the words that are generated according to a language model.
\\
\indent To this end, we propose a novel attention mechanism that is equipped with a component named \textbf{attention sentinel}, on which a decoder can fall back when it chooses not to attend to the source sentence. Hereafter, the improved attention mechanism is referred to as \textbf{adaptive attention} since it can choose adaptively between relying on a regular attention component and falling back on the attention sentinel. We build a new NMT system by integrating an adaptive attention model into the NMT system described in~\cite{Bahdanauetal:15}. To show the effectiveness of adaptive attention, we conduct experiments on Chinese-English translation tasks with standard NIST datasets. Results show that the proposed adaptive attention model achieves an improvement of 0.8 BLEU score. To the best of our knowledge, the adaptive attention method discussed here has not been used before for NMT, although the problem we intend to attack is not new~\cite{Tuetal:16b}.
\\
\indent The rest of the paper is structured as follows. In Section~\ref{approach} we first introduce a conventional attention mechanism for NMT (Section~\ref{background}), then present a detailed description of adaptive attention (Section~\ref{adaptive_attention}). Experiments are presented in Section~\ref{experiments}. After comparing with related work in Section~\ref{related_work}, we finally conclude our work in Section~\ref{conclusion}.

\section{Method}
\label{approach}
\subsection{Attention-based NMT}
\label{background}
We start by describing an NMT model, which builds on the base of an RNN encoder-decoder framework~\cite{Sutskeveretal:14} and attention mechanisms. Given a source sentence $X=\{x_1,\ldots,x_J\}$ and the corresponding target sentence $Y=\{y_1,\ldots,y_K\}$, the model seeks to maximize an objective function that is defined as log-likelihood:
\begin{equation}
\theta^\ast = \argmax \sum_{(X,Y)}logP(Y|X;\theta)
\end{equation}
where $\theta$ are the parameters of the model.
\\
\indent Regarding the decoding phase, the model produces a translation by choosing a target word $y_i$ at each time step $i$.~\footnote{This greedy search process can be extended with beam search in a straightforward way.}  The probability of word prediction is conditioned on the source sentence $X$ and previously generated words $y_i,\ldots, y_{i-1}$,  as defined as the following softmax function:
\begin{equation}
P(y_i|y_{<i},X) = softmax(g(y_{i-1},t_i,c_i))
\end{equation}
where $g(\cdot)$ is a non-linear function, and $t_i$ is a decoding state for the time step $i$, which is initialized with an encoding vector of $X$ and is computed by
\begin{equation}
t_i = f(t_{i-1},x_{i-1},c_i)
\end{equation}
The activation function $f(\cdot)$ can be a vanilla RNN~\cite{Boden:02} or sophisticated units such as Gated Recurrent Unit (GRU)~\cite{Choetal:14} and Long Short-Term Memory (LSTM)~\cite{HochreiterSchmidhuber:97}. For the model in this paper we choose to use GRU. 
\\
\indent $c_i$ in Eq. 2 and Eq. 3 are the \textbf{attention model} from the source sentence, which can be defined as:
\begin{equation}
c_i = \sum_{j=1}^{J}\alpha_{i,j} \cdot h_j
\end{equation}
where $h_j=[\overrightarrow{h}_j^T;\overleftarrow{h}_j^T]^T$ represents the annotation vector of the source word $x_j$ generated by a bi-directional RNN~\cite{SchusterPaliwal:97}, and the weight $\alpha_{i,j}$ is calculated as follows:
\begin{equation}
\alpha_{i,j} = \frac{e_{i,j}}{\sum_{j=1}^{J}e_{i,j}}
\end{equation}
Here $e_{i,j}$ measures the similarity between the target word $y_i$ and the source word $x_j$, which can be calculated with diverse methods~\cite{Luongetal:15b}. In this paper we specifically utilize the following one:
\begin{equation}
\begin{array} {lcl} e_{i,j} & = & a(t_{i-1},h_j) \\ & = & V_a^T tanh(W_a t_{i-1}+U_a h_j) \end{array}
\end{equation}
where $V_a^T$, $W_a$, and $U_a$ are parameters to be learned.
The architecture of the attention model described above is depicted in Figure~\ref{attention}(a).

\subsection{Adaptive Attention Model}
\begin{figure}[tb]
\small
\begin{center}
\begin{tabular}{cc}
 \multicolumn{2}{l}{\includegraphics[scale=0.4]{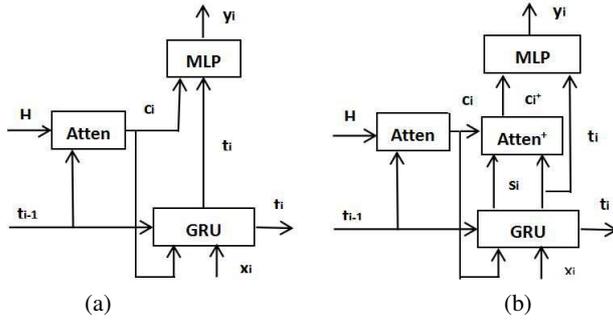}} \\
\multicolumn{2}{c}{(a) \hspace{5cm} (b)} \\
\end{tabular}
\end{center}
\caption{The architecture of conventional attention model (a) and adaptive attention model (b).} \label{attention}
\end{figure}

\noindent Although the attention model introduced above has shown its effectiveness in NMT, it cannot tell when a decoder should use the attention model information and when the decoder should not. Motivated from the work in~\cite{Merityetal:16,Luetal:16}, we introduce the concept of \textbf{attention sentinel}, which is a latent representation of what a decoder already knows. A decoder can fall back on the attention sentinel when it chooses to ``omit" the source sentence for some time steps.
\\
\indent \textbf{Attention sentinel}: A decoder's memory stores information from both the source sentence and the target-size language model. From the state we learn a new component that can be used when the decoder chooses not to attend to the source sentence. Such a component is called the attention sentinel. For a decoder that uses GRU-RNN, the attention sentinel vector $s_i$ is defined by Eq. 7 and Eq. 8.~\footnote{The attention sentinel for LSTM-RNN can be defined in a similar way; Readers interested can refer to~\cite{Luetal:16} for a detailed description.}
\begin{equation}
g_i = \sigma(W_xx_i+W_tt_{i-1})
\end{equation}
\begin{equation}
s_i = g_i \odot tanh(W_st_i)
\end{equation}
where $W_x$, $W_t$, and $W_s$ are parameters of the attention sentiel, $x_i$ is the input to GRU at the time step $i$, $\sigma$ represents a sigmoid activation function, and $\odot$ means an element-wise product.
\\
\indent Based on the attention sentinel, we can propose our adaptive attention model (depicted in Figure~\ref{attention}(b)). The new model has a context vector $c_i^+$, defined as a linear combination of $s_i$ and $c_i$:
\begin{equation}
c_i^+ = \beta_is_i + (1-\beta)c_i
\end{equation}
where $\beta_i$ is a \textbf{sentinel gate} at time step $i$, which always takes a scalar value between 0 and 1. A value of 1 means that only attention sentinel information is used.
To learn the parameter $\beta_i$, we extend the vector $e_i=[e_{i1},\ldots,e_{iJ}]$ with a new element
\begin{equation}
\hat{e_i}  =  [e_i;W_h^Ttanh(W_ss_i + U_ah_i)]
\end{equation}
where $[.;.]$ indicates vector concatenation, and $W_s$ and $W_h$ are parameters to be learned.
Now the weights for the adaptive attention model are computed by:
\begin{equation}
\hat{\alpha}_i  = softmax(\hat{e}_i) \\
\end{equation}
Here $\hat{\alpha}_i$ is a vector with $J+1$ dimensions. We take the last element of the vector as the sentinel gate value: $\beta_i=\hat{\alpha}_i[J+1]$.
\\
\indent \textbf{Decoder prediction}: The prediction over a vocabulary is a standard softmax function with an extended attention mechanism:
\begin{equation}
p_i = softmax(W_p(c_i^++t_i))
\end{equation}
where $W_p$ are parameter to be learned.
\label{adaptive_attention}

\section{Experiments}
\label{experiments}
\subsection{Setup}
We conducted experiments on NIST Chinese-English translation tasks. Our training data consists of 1.25M sentence pairs extracted from LDC corpora,~\footnote{The corpora include LDC2002E18, LDC2003E07, LDC2003E14, Hansards portion of LDC2004T07, LDC2004T08, and LDC2005T06.} which contain 27.9M Chinese words and 34.5M English words, respectively. In all the experiments, we used the NIST 2006 dataset (1,664 sentence pairs) for system development and tested the system on the NIST 2003, 2004, 2005 datasets (919, 1,788, 1,082 sentence pairs, respectively). We used the case-insensitive 4-gram NIST BLEU score~\cite{Papinenietal:02} as the evaluation metric.
\\
\indent For efficient training of neural networks, we utilized sentences of length up to 50 words in the training data. Moreover, we limited the source and target vocabularies to the 16K most frequent words, which cover $95.8\%$ and $98.2\%$ word tokens of source and target sentences, respectively. All the out-of-vocabulary (OOV) words were mapped to the special token {\em UNK}. The word embedding dimension is set to 620 and the size of a hidden layer is set to 1,000. The beam size for translation is set to 10. All the other settings are the same as in~\cite{Bahdanauetal:15}.
\\
\indent We compared our system with two representative translation systems, one for conventional statistical machine translation (SMT) and the other for NMT.
\begin{itemize}
\item \textbf{cdec}~\cite{Dyeretal:10}: an open-source hierarchical phrase-based translation system~\cite{Chiang:07} with default configuration and 4-gram language model trained on the target sentences of training data.~\footnote{\url{https://github.com/redpony/cdec}}
    \item \textbf{RNNSearch}: a re-implementation of the attention-based neural machine translation system~\cite{Bahdanauetal:15} with slight changes from dl2mt tutorial.~\footnote{\url{https://github.com/nyu-dl/dl4mt-tutorial}} RNNSearch uses GRU as the activation function of an RNN and incorporates dropout~\cite{Srivastavaetal:12} on the output layer. We use AdaDelta~\cite{Zeiler:12} to optimize model parameters. For translation, the beam size is also set to 10.
\end{itemize}

\subsection{Main Results}
\begin{table*}[tb]
\begin{center}
\begin{tabular}{c|c|c|c|c|c|c|c|c}
\hline
\# & System & \#Params & Time & NIST06 & NIST03 & NIST04 & NIST05 & All$^\ast$ \\
\hline
1 & cdec & - & - & 33.4 & 33.0 & 35.7 & 32.1 & 34.2 \\
\hline
2 & RNNSearch & 60.6M & 153m & 34.0 & 33.7 & 37.0 & 34.1 & 35.4 \\
\hline
3 & Adaptive & 70.6M & 207m & 34.7$^\dagger$ & 34.2 & 38.0$^\ddagger$ & 35.0$^\dagger$ & 36.2$^\ddagger$ \\
\hline
\end{tabular}
\caption{Main results on the datasets of NIST 2003, 2004, 2005, with NIST 2006 as the development set. $^\ast$The results were achieved on a union set of NIST 2003, 2004, and 2005 data. $^\dagger$ and $^\ddagger$: significant over RNNSearch at the level of 0.05 and 0.01, respectively, as tested by bootstrap resampling~\cite{Koehn:04}.} \label{main_results}
\end{center}
\end{table*}
The main results are shown in Table~\ref{main_results}, where the parameter size, training speed, and performance of each system are presented. From the results we can see that the best adaptive attention model achieved an improvement of 0.7 BLEU score over RNNSearch on the development set. We then evaluated the same model on the test sets, and a significant improvement of 0.8 BLEU score was achieved over RNNSearch (the improvement over cdec is 2.0 BLEU score). On the other hand, we find that adaptive attention model incurs more parameters than RNNSearch (60.6M vs. 70.6M). And more training time is required (153minutes/epoch vs. 207minutes/epoch).
\subsection{Analysis}
It is interesting to examine what kind of words tends not to attend to source sentences. To this end, we translated the set of Chinese sentences extracted from NIST 2003, 2004, and 2005 datasets, and recorded all the predicted target words that have a sentinel gate value greater than or equal to 0.9. From the resulted word list, we present the top 15 most frequent words and their frequency counts in Table~\ref{top_beta_words}.
\begin{table}[tb]
\begin{center}
\small \begin{tabular}{c|c|c|c|c|c}
word & \# & word & \# & word & \#   \\
\hline
the & 6,595 & UNK & 4,507 & to & 1,274  \\
\hline
a & 1,112 & of & 828 & that  & 810 \\
\hline
and & 770 & - & 667 & be & 542 \\
\hline
`` & 467 & year & 418 & states & 347 \\
\hline
us & 331 & not & 294 & been & 285 \\
\end{tabular}
\end{center}
\caption{The top 15 most frequent words from testing data that are assigned a sentinel gate value $\geq 0.9$.}\label{top_beta_words}
\end{table}
\\
\indent From the table we can see that the translation system is inclined to rely on the attention sentinel to generate auxiliary words, such as {\em the} and {\em to}. This observation is consistent with our intuition. Regarding the token {\em UNK}, recall that the symbol is a mapping from OOV words, whose lexical information is lost due to the mapping. Thus, resorting to the attention sentinel to predict {\em UNK} is an appropriate choice. Finally, {\em states} appears in the top word list because  this word, most of the time, occurs immediately after the word {\em united} in our data. Thus, {\em states} can be predicted without referring to the source sentence when {\em united} appears as the preceding word. Inspired by the observation, we further conclude that the adaptive attention model can help predict words in named entities and collocations, in addition to unaligned words.
\section{Related Work}
\label{related_work}
Attention mechanism have become a standard component of state-of-the-art neural NMT systems in some sense. Bahdanau et al.~\shortcite{Bahdanauetal:15} propose a model to jointly align and translate words. Luong et al.~\shortcite{Luongetal:15b} propose and compare diverse attention models. Tu et al.~\shortcite{Tuetal:16a} propose to extend attention models with a coverage vector in order to attack the problem of under-translation and over-translation. All the previous attention models work well, but they cannot tell when not to attend to source sentences.
\\
\indent Our work is inspired by Lu et al.~\shortcite{Luetal:16}, which propose an adaptive attention model for the task of image captioning. The main difference is that they build their adaptive attention model on the base of a spatial attention model, which is different from conventional attention models for NMT. Moreover, our adaptive attention model uses GRU as the RNN activation function while Lu et al.~\cite{Luetal:16} adopt LSTM. Regarding the literature of NMT, the most related work is Tu et al.~\shortcite{Tuetal:16b}, which utilize a context gate to trade off the source-side and target-side context. In this paper, we instead focus on designing a new attention mechanism.
\section{Conclusion}
\label{conclusion}
In this paper, we addressed the problem of learning when to attend to source sentence. We introduced a new component named attention sentinel, based which we built an adaptive attention model. Experiments on NIST Chinese-English translation tasks show that the model achieved a significant improvement of 0.8 BLEU score.

\bibliography{acl2017}

\begin{thebibliography}{}
\expandafter\ifx\csname natexlab\endcsname\relax\def\natexlab#1{#1}\fi

\bibitem[{Bahdanau et~al.(2015)Bahdanau, Cho, and Bengio}]{Bahdanauetal:15}
Dzmitry Bahdanau, Kyunghyun Cho, and Yoshua Bengio. 2015.
\newblock Neural machine translation by jointly learning to align and
  translate.
\newblock In {\em Proceedings of ICLR\/}.

\bibitem[{Boden(2002)}]{Boden:02}
Mikael Boden. 2002.
\newblock A guide to recurrent neural networks and back-propagation.
\newblock In {\em the Dallas project\/}.

\bibitem[{Chiang(2007)}]{Chiang:07}
David Chiang. 2007.
\newblock Hierarchical phrase-based translation.
\newblock {\em Computational Linguistics\/} 33(2):201--228.

\bibitem[{Cho et~al.(2014)Cho, Merrienboer, Gulcehre, Bougares, Schwenk, and
  Bengio}]{Choetal:14}
Kyunghyun Cho, Bart~van Merrienboer, Caglar Gulcehre, Fethi Bougares, Holger
  Schwenk, and Yoshua Bengio. 2014.
\newblock Learning phrase representations using rnn encoder-decoder for
  statistical machine translation.
\newblock In {\em Proceedings of EMNLP\/}. pages 1724--1734.

\bibitem[{Dyer et~al.(2010)Dyer, Lopz, Ganitkevitch, Weese, Ture, Blunsom,
  Setiawan, Eidelman, and Resnik}]{Dyeretal:10}
Chris Dyer, Adam Lopz, Juri Ganitkevitch, Jonathan Weese, Ferhan Ture, Phil
  Blunsom, Hendra Setiawan, Vladimir Eidelman, and Philip Resnik. 2010.
\newblock cdec: A decoder, alignment, and learning framework for finite-state
  and context-free translation models.
\newblock In {\em Proceedings of ACL 2010 System Demonstrations\/}. pages
  7--12.

\bibitem[{Hochreiter and Schmidhuber(1997)}]{HochreiterSchmidhuber:97}
Sepp Hochreiter and J$\ddot{u}$rgen Schmidhuber. 1997.
\newblock Long short-term memory.
\newblock {\em Neural Computation\/} 9(8):1735--1780.

\bibitem[{Koehn(2014)}]{Koehn:04}
Philipp Koehn. 2014.
\newblock Statistical significance tests for machine translation evaluation.
\newblock In {\em Proceedings of EMNLP\/}. pages 388--395.

\bibitem[{Lu et~al.(2016)Lu, Xiong, Parikh, and Richard}]{Luetal:16}
Jiasen Lu, Caiming Xiong, Devi Parikh, and Socher Richard. 2016.
\newblock Knowing when to look: Adaptive attention via a visual sentinel for
  image captioning.
\newblock In {\em arXiv preprint arxiv:1612.01887\/}.

\bibitem[{Luong et~al.(2015{\natexlab{a}})Luong, Le, Vinyals, and
  Zaremba}]{Luongetal:15a}
Minh-Thang Luong, Quoc~V. Le, Oriol Vinyals, and Wojciech Zaremba.
  2015{\natexlab{a}}.
\newblock Addressing the rare word problem in neural machine translation.
\newblock In {\em Proceedings of ACL\/}. pages 11--19.

\bibitem[{Luong et~al.(2015{\natexlab{b}})Luong, Pham, and
  Manning}]{Luongetal:15b}
Minh-Thang Luong, Hieu Pham, and Christopher~D. Manning. 2015{\natexlab{b}}.
\newblock Effective approaches to attention-based neural machine translation.
\newblock In {\em Proceedings of EMNLP\/}. pages 1412--1421.

\bibitem[{Merity et~al.(2016)Merity, Xiong, Bradbury, and
  Richard}]{Merityetal:16}
Stephen Merity, Caiming Xiong, James Bradbury, and Socher Richard. 2016.
\newblock Pointer sentinel mixture models.
\newblock In {\em arXiv preprint arXiv:1609.07843\/}.

\bibitem[{Papineni et~al.(2002)Papineni, Roukos, Ward, and
  Zhu}]{Papinenietal:02}
Kishore Papineni, Salim Roukos, Todd Ward, and Wei-Jing Zhu. 2002.
\newblock Bleu: A method for automatic evaluation.
\newblock In {\em Proceedings of ACL\/}. pages 311--318.

\bibitem[{Schuster and Paliwal(1997)}]{SchusterPaliwal:97}
Mike Schuster and Kuldip~K Paliwal. 1997.
\newblock Bidirectional recurrent neural networks.
\newblock {\em IEEE Transactions on Signal Processing\/} 45(1):2673--2681.

\bibitem[{Srivastava et~al.(2012)Srivastava, Krizhevsky, Sutskever,
  Salakhutdinov, and Hinto}]{Srivastavaetal:12}
Nitish Srivastava, Alex Krizhevsky, Ilya Sutskever, Ruslan~R. Salakhutdinov,
  and Geoffrey~E. Hinto. 2012.
\newblock Improving neural networks by preventing co-adaptation of feature
  detectors.
\newblock In {\em arXiv preprint arXiv:1207.0580\/}.

\bibitem[{Sutskever et~al.(2014)Sutskever, Vinyals, and Le}]{Sutskeveretal:14}
Ilya Sutskever, Oriol Vinyals, and Quoc~V Le. 2014.
\newblock Sequence to sequence learning with neural networks.
\newblock In {\em Proceedings of NIPS\/}.

\bibitem[{Tu et~al.(2016{\natexlab{a}})Tu, Liu, Lu, Liu, and Li}]{Tuetal:16b}
Zhaopeng Tu, Yang Liu, Zhengdong Lu, Xiaohua Liu, and Hang Li.
  2016{\natexlab{a}}.
\newblock Context gates for neural machine translation.
\newblock {\em TACL\/} 5:87--99.

\bibitem[{Tu et~al.(2016{\natexlab{b}})Tu, Lu, Liu, Liu, and Li}]{Tuetal:16a}
Zhaopeng Tu, Zhengdong Lu, Yang Liu, Xiaohua Liu, and Hang Li.
  2016{\natexlab{b}}.
\newblock Modeling coverage for neural machine translation.
\newblock In {\em Proceedings of ACL\/}. pages 76--85.

\bibitem[{Yang and Maosong(2015)}]{LiuSun:15}
Liu Yang and Sun Maosong. 2015.
\newblock Contrastive unsupervised word alignment with non-local features.
\newblock In {\em Proceedings of AAAI\/}. pages 857--868.

\bibitem[{Zeiler(2012)}]{Zeiler:12}
Matthew~D. Zeiler. 2012.
\newblock Adadelta: An adaptive learning rate method.
\newblock In {\em arXiv preprint arXiv:1212.5701\/}.

\end{thebibliography}
\bibliographystyle{acl_natbib}

\end{CJK}
\end{document}